\documentclass[runningheads]{llncs}

\usepackage{amsmath}
\usepackage{amssymb}
\usepackage{graphicx}
\usepackage{algorithm}
\usepackage{algpseudocode}
\usepackage{bbm}
\usepackage{multirow,multicol}
\usepackage{mathtools}
\usepackage{CJKutf8}
\usepackage{footnote}
\usepackage{url}
\usepackage{makecell}

\usepackage{pifont}% http://ctan.org/pkg/pifont
\newcommand{\cmark}{\ding{51}}%
\newcommand{\xmark}{\ding{55}}%

\algrenewcommand{\algorithmicrequire}{\textbf{Input:}}
\algrenewcommand{\algorithmicensure}{\textbf{Output:}}
\algnewcommand\AND{\textbf{and} }

\makeatletter
\newcommand{\printfnsymbol}[1]{%
  \textsuperscript{\@fnsymbol{#1}}%
}
\makeatother

\begin{document}
\begin{CJK*}{UTF8}{gbsn}

%\title{Robust Spoken Language Understanding with Value Refinement and Reinforcement Learning}
\title{Robust Spoken Language Understanding with RL-based Value Error Recovery}
%\title{Reinforcement Learning for Robust Spoken Language Understanding with Rule-based \\ Value Refinement}
\titlerunning{Robust SLU with RL-based Value Error Recovery}

\author{Chen Liu\thanks{Chen Liu and Su Zhu contributed equally to this work.} \and Su Zhu\printfnsymbol{1}  \and Lu Chen \and Kai Yu\thanks{Lu Chen and Kai Yu are the corresponding authors.}}
\authorrunning{Liu et al.}

\institute{
MoE Key Lab of Artificial Intelligence, AI Institute, Shanghai Jiao Tong University \\
SpeechLab, Department of Computer Science and Engineering \\
Shanghai Jiao Tong University, Shanghai, China \\
\email{\{chris-chen,paul2204,chenlusz,kai.yu\}@sjtu.edu.cn}
}

\maketitle              % typeset the header of the contribution
\begin{abstract}
Spoken Language Understanding (SLU) aims to extract structured semantic representations (e.g., \emph{slot-value} pairs) from speech recognized texts, which suffers from errors of Automatic Speech Recognition (ASR). To alleviate the problem caused by ASR-errors, previous works may apply input adaptations to the speech recognized texts, or correct ASR errors in predicted values by searching the most similar candidates in pronunciation. However, these two methods are applied separately and independently. In this work, we propose a new robust SLU framework to guide the SLU input adaptation with a rule-based value error recovery module. The framework consists of a slot tagging model and a rule-based value error recovery module. We pursue on an adapted slot tagging model which can extract potential slot-value pairs mentioned in ASR hypotheses and is suitable for the existing value error recovery module. After the value error recovery, we can achieve a supervision signal (reward) by comparing refined slot-value pairs with annotations. Since operations of the value error recovery are non-differentiable, we exploit policy gradient based Reinforcement Learning (RL) to optimize the SLU model. Extensive experiments on the public CATSLU dataset show the effectiveness of our proposed approach, which can improve the robustness of SLU and outperform the baselines by significant margins.

\end{abstract}

\section{Introduction}
\label{sec:intro}

The Spoken Language Understanding (SLU) module is a key component of Spoken Dialogue System (SDS), parsing user's utterances into structured semantic forms. For example, ``\emph{I want to go to Suzhou not Shanghai}'' can be parsed into ``\emph{\{inform(dest=Suzhou), deny(dest=Shanghai)\}}''. It can be usually formulated as a sequence labelling problem to extract values (e.g., \emph{Suzhou} and \emph{Shanghai}) for certain semantic slots (attributes, e.g., \emph{inform-dest} and \emph{deny-dest}).

%Slot filling can be done with \emph{aligned} or \emph{unaligned} data, where user utterances are annotated at word level and sentence level, respectively.

\begin{figure}[t]
    \centering
    \includegraphics*[width=0.8\textwidth]{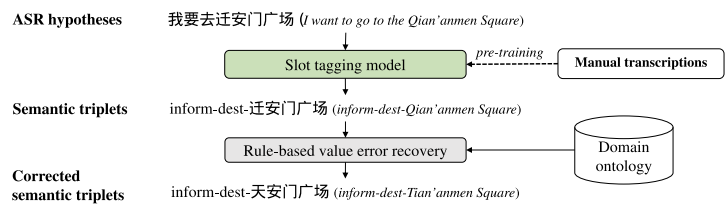}
    \caption{An overview of the robust SLU framework, which is composed of two main components: a slot tagging model and a rule-based value error recovery module. During evaluation, only ASR hypotheses are fed into the two modules to generate the final semantic form.} %
    \label{fig:model_arch}
\end{figure}

It is crucial for SLU to be robust to speech recognition errors, since ASR-errors would be propagated to the downstream SLU model. By ignoring ASR-errors, it promotes rapid development of natural language processing (NLP) algorithms for SLU~\cite{mesnil2013investigation,yao2014spoken,liu2016attention,goo2018slot} where SLU models are trained and evaluated on manual transcriptions and even natural language texts. Once ASR hypotheses are used as input for evaluation, it will lead to a sharp decrease in SLU performance~\cite{zhu2019catslu}.

%Since ASR module which converts audio to text would take some recognition errors due to complex noise environments and domain specific expressions

ASR-errors may give rise to two issues: 1) inputs for training and evaluation are mismatched; 2) Sequence labelling models extract values directly from ASR hypotheses, which may contain wrong words. Previous works try to overcome these problems in two ways: 1) Adaptive training approaches are introduced to transfer the SLU model trained on manual transcriptions to ASR hypotheses~\cite{zhu2018robust,schumann2018incorporating}. 2) Other works adopt rule-based post-processing techniques to refine the predicted values with the most similar candidates in pronunciation~\cite{tan2019multi,wang2019transfer,li2019robust}. However, this value error recovery module is usually fixed and independent of the SLU model.

To overcome the above problems, we propose a new robust SLU framework to guide the former SLU model trained with a rule-based value error recovery. As illustrated in Figure \ref{fig:model_arch}, it consists of a slot tagging model and a value error recovery module. The slot tagging model is pre-trained on manual transcriptions, which considers SLU as a sequence labelling problem. To alleviate the input mismatched issue, it is adaptively trained on ASR hypotheses. The value error recovery module is exploited to correct potential ASR-errors in predicted values of the slot tagging model, which is built upon a pre-defined domain ontology.

However, there are no word-aligned annotations for ASR hypotheses to finetune the slot tagging model. Thus, we indirectly guide the adaptive training of the slot tagging model on ASR hypotheses by utilizing supervisions of the value error recovery. Concretely, we can compute a reward by measuring predicted semantic forms after the value error recovery with annotations, and then optimize the slot tagging model by maximizing the expected reward. Since operations in the value error recovery are non-differentiable, we use a policy gradient~\cite{sutton2000policy} based reinforcement learning (RL) approach for optimizing.

We conduct an empirical study of our proposed method and a set of carefully selected state-of-the-art baselines on the 1st CATSLU challenge dataset~\cite{zhu2019catslu}, which is a large-scale Chinese SLU dataset collected from a real-world SDS application. Experiment results confirm that our proposed method can outperform the baselines significantly.

In summary, this paper makes the following contributions:
\begin{itemize}
    \item To the best of our knowledge, this is the first work to train a slot tagging model guided by a rule-based value error recovery module. It tends to learn a robust slot tagging model for easier and more accurate value error recovery.
    \item We propose to optimize the slot tagging model with indirect supervision and RL approach, which does not require word-aligned annotations on ASR hypotheses. Ablation study confirms that RL training can give improvements even without the value error recovery module.
\end{itemize}

\section{Proposed Method}
\label{sec:method}

In this section, we provide details of our proposed robust SLU framework, which consists of a sequence labelling based slot tagging model and a rule-based value error recovery (VER) module. To guide the training of the slot tagging model on ASR hypotheses with the VER, we propose an RL-based training algorithm. 

Let $x=(x_1 \cdots x_{|x|})$ and $\hat{x}=(\hat{x}_1 \cdots \hat{x}_{|\hat{x}|})$ denote the ASR 1-best hypothesis and manual transcription of one utterance respectively. Its semantic representation (i.e., \emph{act(slot=value)} triplets) is annotated on $\hat{x}$. Thus, it is easy to get the word-level tags on $\hat{x}$, $\hat{o}=(\hat{o}_1 \cdots \hat{o}_{|\hat{x}|})$, which is in Begin/In/Out (BIO) schema (e.g., $O$, $B$-inform-dest, $B$-deny-dest), as shown in Table \ref{tab:data_example}.

\begin{table}[t]
    \caption{An example of user utterance (manual transcription and ASR hypothesis) and its semantic annotations.}
    \label{tab:data_example}
    \small
    \centering
    \begin{tabular}{p{0.5cm}|l}
        \hline
        $\hat{x}$ & I want to go to Suzhou not Shanghai \\
         \hline
        $x$ & I one goal to Suizhou not Shanghai \\
         \hline
        $y$ & inform(dest=``Suzhou''); deny(dest=``Shanghai'') \\
         \hline
        $\hat{o}$ & I$_{[O]}$ want$_{[O]}$ to$_{[O]}$ go$_{[O]}$ to$_{[O]}$ Suzhou$_{[B\text{-inform-dest}]}$ not$_{[O]}$ Shanghai$_{[B\text{-deny-dest}]}$ \\
         \hline
    \end{tabular}
\end{table}

% \footnote{We convert unaligned labels into aligned labels by combining the \emph{act} and \emph{slot} as a whole tag. For instance, the tags of the example given in Sec. \ref{sec:intro} can be ``\emph{go:O to:O Beijing:B-inform-dest not:O Shanghai:B-deny-dest}''.} 

% 2.1.1
\subsection{Slot Tagging Model}
\label{subsec:basic_model}

For slot tagging, we adopt an encoder-decoder model with focus mechanism~\cite{zhu2017encoder} to model the label dependency. A BLSTM encoder reads an input sequence $\hat{x}$, and generates the hidden states at the $t$-th time-step via 
\begin{align}
\mathbf{h}_t=[\overrightarrow{\mathbf{h}_t}; \overleftarrow{\mathbf{h}_t}];~\overrightarrow{\mathbf{h}_t}=\operatorname{LSTM_f}(\overrightarrow{\mathbf{h}_{t-1}}, \phi(\hat{x}_t));~\overleftarrow{\mathbf{h}_t}=\operatorname{LSTM_b}(\overleftarrow{\mathbf{h}_{t+1}}, \phi(\hat{x}_t))
\end{align}
where $\phi(\cdot)$ is a word embedding function, $[\cdot;\cdot]$ denotes vector concatenation, $\operatorname{LSTM_f}$ and $\operatorname{LSTM_b}$ represent the forward and backward LSTMs, respectively.

Then, an LSTM decoder updates its hidden states at the $t$-th time-step recursively by $\mathbf{s}_t=\operatorname{LSTM}(\mathbf{s}_{t-1}, [\psi(\hat{o}_{t-1}); \mathbf{h}_t])$, where $\psi(\cdot)$ is a label embedding function, and $\mathbf{s}_0$ is initialized with $\overleftarrow{\mathbf{h}}_1$. Finally, the slot tag $\hat{o}_t$ is generated by
\begin{align}
P(\hat{o}_t|\hat{o}_{<t};\hat{x})=\operatorname{Softmax}(\mathbf{W}\mathbf{s}_t+\mathbf{b})
\end{align}
where $\mathbf{W}$ and $\mathbf{b}$ are parameters for the linear output layer.

Following the BIO schema, we can restructure the predicted slot-tag sequence aligned with the input sequence to obtain a set of \emph{act(slot=value)} triplets.

% 2.1.2
\subsection{Value Error Recovery (VER) Module}

During evaluation, ASR hypotheses are fed into the slot tagging model to get the \emph{act(slot=value)} triplets, which may retain ASR errors in the values. Thus, a value error recovery (VER) module based on a pre-defined domain ontology~\footnote{All possible value candidates of each slot are provided in the domain ontology.} is applied to refine wrong values. Li et al.~\cite{li2019robust} search through the ontology to find the most similar candidate with minimum edit distance. While the  calculation of the minimum edit distance is time-consuming and hard to be parallelized, we exploit an N-gram based cosine distance to accelerate this process.

Generally, we define the $n$-gram set of a word sequence $w=(w_1, \cdots, w_T)$ as $\operatorname{Ngram}(w, n)=\{(w_i,\cdots,w_{i+n-1})\;|\;i=\{1, \cdots, T-n+1\}\}$. The $n$-grams of all values in the domain ontology $\mathcal{O}$ constitute a vocabulary, denoted as $\operatorname{Ngram}(\mathcal{O}, n)$. Given a predicted semantic triplet $a(s=v)$, where the value $v$ is a word sequence $v=(v_1,\cdots,v_T)$. Then, we get a binary-valued feature vector $\mathbf{d}'(v)=(d_1^v,\cdots,d_L^v)$ for the predicted value $v$, where $L=|\operatorname{Ngram}(\mathcal{O}, n)|$ and $d_j^v=\mathbbm{1}_{\{\operatorname{Ngram}(\mathcal{O}, n)_j \in \operatorname{Ngram}(v, n)\}}$. Finally, we normalize it to be a unit vector, $\mathbf{d}(v)=\mathbf{d}'(v)/||\mathbf{d}'(v)||$.

Based on the domain ontology, there is a value candidate set~\footnote{E.g., the value candidate set for slot \emph{address} can be all available addresses saved in the database of a dialogue system.} corresponding to the act $a$ and slot $s$, $\mathcal{V}_{a,s}=(\Bar{v}^1, \cdots, \Bar{v}^M)$, where $M$ is the number of possible values. Therefore, the value candidate set can be represented as an $L \times M$ feature matrix $\mathbf{D}(\mathcal{V}_{a,s})$, the $k$-th column of which is $\mathbf{d}(\Bar{v}^k)$. We believe that the more $n$-grams two values share, the more similar they are. Thus, we calculate the cosine similarity score between $\mathbf{d}(v)$ and each column vector in $\mathbf{D}(\mathcal{V}_{a,s})$ as:
\begin{align}
\textbf{sim}_{\text{word}}(v,\mathcal{V}_{a,s})=\mathbf{D}(\mathcal{V}_{a,s})^\top \mathbf{d}(v), ~\in\mathbb{R}^M
\end{align}

Since ASR tends to produce words similar in pronunciation, we convert word sequences of values into pronunciation sequences with a pre-defined pronunciation dictionary. For example, ``上海~(Shanghai)'' can be converted into ``\emph{sh ang h ai}''. Therefore, we get another similarity vector by considering the pronunciation $n$-grams, denoted as $\textbf{sim}_{\text{pron}}(v,\mathcal{V}_{a,s})$. The final similarity vector is obtained by averaging the two vectors, i.e., 
\begin{equation}
\textbf{sim}(v,\mathcal{V}_{a,s})=\lambda\textbf{sim}_{\text{word}}(v,\mathcal{V}_{a,s})+(1-\lambda)\textbf{sim}_{\text{pron}}(v,\mathcal{V}_{a,s})
\end{equation}
where $\lambda$ is a balancing parameter ($0.5$ in our experiments). So far, we can easily find the best alternative value $\Bar{v}^k$, where $k=\operatorname{argmax}(\textbf{sim}(v,\mathcal{V}_{a,s}))$.

Although some slots have numerous possible values in the domain ontology, it is much efficient by simply performing matrix multiplication. We also set a threshold ($0.5$ in this paper) to reject a bad error recovery.

% 2.3
\subsection{Training Procedure}
\label{subsec:train_proc}

% \begin{algorithm}[t]
% \small
% \caption{Training algorithm}
% \label{alg:train}
% \begin{algorithmic}[1]
% \Require Annotated transcriptions $\mathcal{D}_{tscp}=\{(u, o)\}$; sentence-level labelled hypotheses $\mathcal{D}_{hyp}=\{(r, y)\}$; reward function $R(\cdot)$.
% \Ensure Robust slot tagging model $\Theta$
% \State Initialize $\Theta$ randomly;
% \State Pre-train the model using $\mathcal{D}_{tscp}$ supervised by Eqn.(\ref{eqn:tag_loss});
% \Repeat
% \State Sample $(r, y)$ from $\mathcal{D}_{hyp}$;
% \State $K$ groups of semantic labels $\Tilde{y}^1,...,\Tilde{y}^K$ are generated after beam search and error correction;
% \For{$k=1...K$}
% \State Compute reward $R(r,y,\Tilde{y}^k)$ by Eqn.(\ref{eqn:reward});
% \EndFor
% \State Compute policy gradient $\nabla_\Theta\hat{E}[R]$ by Eqn.(\ref{eqn:pg});
% \State Update the model: $\Theta \gets \Theta +  \eta_1\nabla_\Theta\hat{E}[R]$;
% \State Sample $(u, o)$ from $\mathcal{D}_{tscp}$;
% \State Update the model: $\Theta \gets \Theta - \eta_2\nabla_\Theta L_{tag}(\Theta)$;
% \Until \text{convergence}
% \end{algorithmic}
% \end{algorithm}

\begin{algorithm}[t]
\small
\caption{Training algorithm}
\label{alg:train}
\begin{algorithmic}[1]
\Require Manual transcriptions with word-aligned labels $\mathcal{D}_{tscp}=\{(\hat{x}, \hat{o})\}$; ASR hypotheses with utterance-level labels $\mathcal{D}_{hyp}=\{(x, y)\}$; reward function $R(\cdot)$.
\Ensure Robust slot tagging model $\Theta$
\State Initialize $\Theta$ randomly;
\Repeat \Comment{Pre-training stage}
\State Sample $(\hat{x}, \hat{o})$ from $\mathcal{D}_{tscp}$;
\State Update the model: $\Theta \gets \Theta - \eta_1\nabla_\Theta L_{tag}(\Theta)$;
\Until \text{convergence}
\Repeat \Comment{RL-training stage}
\State Sample $(x, y)$ from $\mathcal{D}_{hyp}$;
\State $K$ groups of semantic labels $\Tilde{y}^1,...,\Tilde{y}^K$ are generated after beam search and value error recovery by feeding $x$;
\For{$k=1,...,K$}
\State Compute reward $R(x,y,\Tilde{y}^k)$ by Eqn.(\ref{eqn:reward});
\EndFor
\State Compute policy gradient $\nabla_\Theta\hat{E}[R]$ by Eqn.(\ref{eqn:pg});
\State Update the model: $\Theta \gets \Theta +  \eta_2\nabla_\Theta\hat{E}[R]$;
\State Sample $(\hat{x}, \hat{o})$ from $\mathcal{D}_{tscp}$;
\State Update the model: $\Theta \gets \Theta - \eta_1\nabla_\Theta L_{tag}(\Theta)$;
\Until \text{convergence}
\end{algorithmic}
\end{algorithm}

We propose to guide the adaptive training of the slot tagging on ASR hypotheses with the value error recovery module. It takes two advantages: 1) mitigating the input mismatch problem of training and testing; 2) not requiring word-aligned annotations on ASR hypotheses. Meanwhile, it tends to learn a robust slot tagging model suitable for the value error recovery.

We apply the policy gradient based reinforcement learning (RL) algorithm to handle non-differentiable operations. To prune the large search space, the model is pre-trained with annotated transcriptions to bootstrap the RL-training. The whole training procedure contains two stages, as described below.

\subsubsection{Pre-training}

Let $\mathcal{D}_{tscp}=\{(\hat{x}, \hat{o})\}$ denote manual transcriptions with aligned labels. The slot tagging model (let $\Theta$ refer to the model parameters) is trained by minimizing a negative log-likelihood loss:
\begin{align}\label{eqn:tag_loss}
L_{tag}(\Theta) = -\sum_{\mathclap{(\hat{x},\hat{o})\in \mathcal{D}_{tscp}}}\log P(\hat{o}|\hat{x};\Theta).
\end{align}

\subsubsection{RL-training}

ASR hypotheses coupled with unaligned labels, $\mathcal{D}_{hyp}=\{(x, y)\}$, are utilized for the adaptive training. The slot tagging model, $P(o|x;\Theta)$, samples via beam search to produce $K$ tag sequences, and then $K$ sets of \emph{act(slot=value)} triplets. Finally, corrected semantic triplets $\{\Tilde{y}^k\}_{k=1}^K$ are generated after VER module. For each beam, the reward is considered at both triplet-level and utterance-level:
\begin{align}\label{eqn:reward}
R(x, y, \Tilde{y}^k) = R_{\text{triplet}} + R_{\text{utt}} =\left(1-\frac{\text{FP}(y, \Tilde{y}^k)+\text{FN}(y, \Tilde{y}^k)}{|\;y\;|}\right) + \mathbbm{1}_{\{y=\Tilde{y}^k\}}
\end{align}
where the first term punishes false-positives (FP) and false-negatives (FN) of \emph{act(slot=value)} triplets, and the second term is a binary value indicating whether the entire triplets of one utterance is predicted correctly.

The model is optimized by maximizing the expected cumulative rewards using policy gradient descent. The policy gradient can be calculated as:
\begin{align}\label{eqn:pg}
% \begin{split}
% \nabla_\Theta{\hat{E}[R]}&=\frac{1}{K}\sum_{k=1}^K\Big[R(x, y, \Tilde{y}^k) \\ &-B(r)\Big]\cdot\nabla_\Theta\log P(\Tilde{y}^k|r;\Theta)
% \end{split}
\nabla_\Theta{\hat{E}[R]}=\frac{1}{K}\sum_{k=1}^K\Big[R(x, y, \Tilde{y}^k) -B(r)\Big]\cdot\nabla_\Theta\log P(\Tilde{y}^k|x;\Theta)
\end{align}
where $B(r)=\frac{1}{K}\sum_{k=1}^K R(x, y, \Tilde{y}^k)$ is a baseline for reducing the variance of gradient estimation, obtained by averaging the rewards inside a beam.

In order to stabilize the training process, it is beneficial to train batches with $\mathcal{D}_{tscp}$ and $\mathcal{D}_{hyp}$ iteratively. The training framework is shown in Algorithm \ref{alg:train}.

\section{Experiments}
\label{sec:exp}

\subsection{Experimental Setup}

We conduct our experiments on the 1st Chinese Audio-Textual Spoken Language Understanding Challenge (CATSLU)\footnote{\url{https://sites.google.com/view/CATSLU}} dataset containing four dialogue domains (\emph{map}, \emph{music}, \emph{video}, \emph{weather}). The statistics of the CATSLU dataset are demonstrated in detail in Zhu et al.~\cite{zhu2019catslu}.

Slot tagging is modeled at Chinese character level. The $200$-dim char embedding is initialized by pre-training LSTM based bidirectional language models (biLMs) with zhwiki~\footnote{\url{https://dumps.wikimedia.org/zhwiki/latest}} corpus. LSTMs are single-layer with $256$ hidden units. In the training process, parameters are uniformly sampled within the range of $(-0.2, 0.2)$. Dropout with a probability of $0.5$ is applied to non-recurrent layers. We choose Adam~\cite{kingma2014adam} as our optimizer. For the learning rate, we set $\eta_1$=1e-3 and $\eta_2$=5e-4 fixed during training. The maximum norm for gradient clipping is set to $5$. In the RL-training stage, the beam search sampling size $K$ is set to $10$. In the decoding stage, the beam size is $5$. The best model is selected according to the performance on the validation set, and we measure both F$_1$-score of \emph{act(slot=value)} triplets and utterance-level accuracy.

% To leverage domain knowledge, we follow Li et al.~\cite{li2019robust} to add lexicon features as additional input features in both baselines and our proposed model. 

\subsection{Baselines}

We compare the proposed method with strong baselines for robust SLU:

\begin{itemize}
\item \textbf{\textsc{HD}}: Hierarchical Decoding model proposed in Zhao et al.~\cite{zhao2019hierarchical}, which performs slot filling in a generative way with only unaligned data ($\mathcal{D}_{hyp}$).
\item \textbf{\textsc{Focus}}: BLSTM-Focus model as described in section \ref{subsec:basic_model} for slot tagging.
\item \textbf{\textsc{UA}}: Unsupervised Adaptation method~\cite{zhu2018robust} utilizes the language modelling task to transfer the slot tagging model from manual transcriptions to ASR hypotheses.
\item \textbf{\textsc{DA}}: Data Augmentation methods are also involved to predict pseudo labels aligned with ASR hypotheses; thus the pseudo samples can be exploited to train a robust slot tagging model. (1) \textsc{Gen}: ASR hypotheses are fed into the pre-trained slot tagging model to generate pseudo labels. (2) \textsc{Align}: ASR hypotheses are aligned with manual transcriptions via achieving minimum edit-distance, and then the aligned labels of words in transcriptions can be assigned to the corresponding words in ASR hypotheses.
\end{itemize}

\subsection{Main Results}

\begin{table}[t]
    \caption{Main results \emph{with} or \emph{without} lexicon features. F$_1$-score(\%)/joint-accuracy(\%) on the test set of each domain are reported. In the table, \emph{tscp} means manual transcriptions while \emph{hyp} means ASR hypotheses. Our results that significantly outperform the best baseline are marked by $^\dagger$ ($p<0.05$) and $^\ddagger$ ($p<0.01$). $^\star$ denotes oracle experiments, in which manual transcriptions are evaluated.}
    \label{tab:main_res}
    \small
    \centering
    
    \text{(a) \emph{with} lexicon features}
    \begin{tabular}{l|c|c||cccc|c}
        \hline
        \textbf{Models} & \textbf{train} & \textbf{test} & \textbf{map} & \textbf{music} & \textbf{video} & \textbf{weather} & \textbf{avg.}\\
        \hline\hline
        \textsc{HD} & hyp & hyp & 87.8/84.2 & 90.5/82.1 & 88.7/76.1 & 89.6/82.4 & 89.2/81.2 \\
        \hline
        \textsc{Focus}$^\star$ & tscp & tscp & 96.4/93.8 & 97.6/93.3 & 94.1/85.6 & 95.5/90.6 & 95.9/90.8 \\
        \textsc{Focus} & tscp & hyp & 89.0/84.9 & 92.8/84.8 & 91.5/80.9 & 92.6/86.7 & 91.5/84.3 \\
        \hline
        \textsc{UA} & tscp+hyp & hyp & 88.5/85.2 & 91.8/83.6 & 91.2/81.2 & 91.8/84.9 & 90.9/83.7 \\
        \textsc{DA-Gen} & tscp+hyp & hyp & 88.9/85.4 & 92.2/84.6 & 92.0/81.4 & 93.1/87.1 & 91.5/84.6 \\ 
        \textsc{DA-Align} & tscp+hyp & hyp & 89.1/85.5 & 93.1/85.7 & 91.5/80.8 & 93.1/87.0 & 91.7/84.7 \\
        \hline
        Li et al.~\cite{li2019robust} & tscp & hyp & 87.9/83.8 & 92.7/85.1 & \textbf{92.3}/\textbf{82.6} & 93.0/86.8 & 91.5/84.6 \\ 
        \hline
        \multirow{2}{*}{Proposed} & tscp & hyp & 89.0/85.3 & 93.1/85.7 & 91.9/81.6 & 93.1/87.6 & 91.8/85.0 \\
         & tscp+hyp & hyp & \textbf{90.0}/\textbf{86.7}$^\dagger$ & \textbf{93.8}/\textbf{87.0}$^\dagger$ & 92.0/82.0 & \textbf{93.4}/\textbf{87.7} & \textbf{92.3}/\textbf{85.8}$^\dagger$ \\
        \hline
    \end{tabular}
    
    \text{(b) \emph{without} lexicon features}
    \begin{tabular}{l|c|c||cccc|c}
        \hline
        \textbf{Models} & \textbf{train} & \textbf{test} & \textbf{map} & \textbf{music} & \textbf{video} & \textbf{weather} & \textbf{avg.}\\
        \hline\hline
        \textsc{HD} & hyp & hyp & 87.9/83.5 & 87.0/74.1 & 84.0/64.9 & 89.2/80.0 & 87.0/75.6 \\
        \hline
        \textsc{Focus}$^\star$ & tscp & tscp & 96.4/93.9 & 96.3/89.4 & 92.8/81.6 & 94.7/88.7 & 95.0/88.4  \\
        \textsc{Focus} & tscp & hyp & 89.4/86.1 & 92.2/83.6 & 90.2/77.8 & 92.4/85.7 & 91.0/83.3 \\
        \hline
        \textsc{UA} & tscp+hyp & hyp & 89.3/86.4 & 91.9/83.6 & 90.0/78.4 & 91.8/85.0 & 90.8/83.3 \\
        \textsc{DA-Gen} & tscp+hyp & hyp & 88.7/85.7 & 91.4/82.5 & 90.3/78.2 & 92.2/85.8 & 90.6/83.1 \\ 
        \textsc{DA-Align} & tscp+hyp & hyp & 89.3/85.8 & \textbf{92.3}/83.6 &  90.6/77.6 & 91.8/85.1 & 91.0/83.0 \\
        \hline
        \multirow{2}{*}{Proposed} & tscp & hyp & 89.5/86.3 & 91.8/82.8 & 90.8/79.0 & 92.2/86.1 & 91.1/83.6 \\
         & tscp+hyp & hyp & \textbf{89.6}/\textbf{86.8}$^\dagger$ & 92.2/\textbf{83.7} & \textbf{91.2}/\textbf{79.7}$^\ddagger$ & \textbf{92.6}/\textbf{86.2}$^\dagger$ & \textbf{91.4}/\textbf{84.1}$^\ddagger$ \\
        \hline
    \end{tabular}
\end{table}

In this section, the main results on the test set compared with the baselines are demonstrated in Table \ref{tab:main_res}. In the evaluation stage of all baselines and our approach, VER is applied for post-processing. Lexicon features are added as additional input features, same as what Li et al.~\cite{li2019robust} did. 

Overall, the SLU models perform better \emph{with} auxiliary lexicon features. For basic slot filling models, \textsc{Focus} performs much better than HD, showing that the sequence labelling based slot tagging model is more generalizable than a generative model. The results of the oracle experiments suggest that ASR hypotheses largely degrade the performance. By adapting to ASR hypotheses, \textsc{UA} performs slightly better on some domains, but the average result drops instead. With lexicon features, by augmenting the training data with pseudo aligned hypotheses (DA), both \textsc{Gen} and \textsc{Align} can beat the basic model, indicating that \textsc{DA} methods are beneficial for improving robustness to ASR hypotheses. 

With lexicon features, our proposed method outperforms the \textsc{Focus} model in all domains significantly, achieving an average improvement of $0.8\%$ in F$_1$-score and $1.5\%$ in joint-accuracy, which reveals the benefit of VER guided training. Our model also surpasses the best baseline (\textsc{DA-Align}), indicating that it is less effective to merely augment the data with pseudo aligned ASR texts.

We also attempt to employ only transcriptions in training (the second-to-last row), that is, replace $\mathcal{D}_{hyp}$ (on line $7$) with $\mathcal{D}_{tscp}$ in Algorithm \ref{alg:train}. The consistent improvements in all domains compared with \textsc{Focus} prove that the RL loss benefits the slot tagging. On this basis, adaptively involving the ASR hypotheses in training further improves the robustness of the SLU model.

We only compare our model with the ``System 1'' in Li et al.~\cite{li2019robust} (the top solution in CATSLU challenge), because their other systems add the validation set in training and leverage audio information for better results. As shown in the table, our proposed method achieves higher average F$_1$-score and joint-accuracy.

\subsection{Ablation Study}

\subsubsection{Ablation Study of the Slot Tagging Model}

For slot tagging, there are other popular methods like \textsc{BLSTM} and \textsc{BLSTM-CRF}~\cite{li2019robust}. Table \ref{tab:ablation_tag_model} shows the comparison of different slot tagging models. Vanilla \textsc{BLSTM} performs the worst without modeling label dependencies. \text{Focus} can achieve the best results in most cases, thus we choose \textsc{Focus} as the backbone model. It should be noted that our proposed framework can be applied to other slot tagging models.

% \begin{itemize}
% \item \textsc{BLSTM}: The hidden state at each time step is utilized to predict slot tags independently by a linear output layer and the \emph{softmax} function.
% \item \textsc{BLSTM-CRF}: A CRF (conditional random field) layer is placed onto the top of the output layer to model the slot tag dependencies.
% \end{itemize}

\begin{table}[t]
    \small
    \centering
    \caption{Ablation experiments of the slot tagging models. Average F$_1$-score(\%)/joint-accuracy(\%) on the test set are reported with or without lexicon features.}
    \label{tab:ablation_tag_model}
    \begin{tabular}{l||c|c}
        \hline
        \multirow{2}{*}{\textbf{Slot tagging models}} & \multicolumn{2}{c}{\textbf{lexicon features}} \\
        \cline{2-3}
         & \cmark & \xmark \\
        \hline\hline
        \textsc{BLSTM} & 90.84/83.92 & 89.34/80.78 \\
        \textsc{BLSTM-CRF} & 91.31/\textbf{84.49} & 90.30/82.43 \\
        \textsc{Focus} & \textbf{91.46}/84.31 & \textbf{91.03}/\textbf{83.29} \\
        \hline
    \end{tabular}
    \vspace{-0.3cm}
\end{table}

\subsubsection{Effect of the Value Error Recovery (VER) Module}

We apply different post-processing ways for values to examine the effect of the VER module, as shown in Table \ref{tab:ablation_vr}. For \textsc{HD}, \textsc{Focus} and \textsc{DA-Align}, only the evaluation stage is affected by the post-processing.  Results show that it is beneficial to delete invalid triplets (i.e., out of the domain ontology) while finding proper value alternative via the VER module brings further improvement. For our proposed method, both training and evaluation stages involve the VER module. Our proposed method makes consistent improvement regardless of the post-processing ways, whereas VER works best. Due to the additional triplet- and utterance-level policy losses which help adapt the tagging model to ASR hypotheses, improvements are still observed even without the post-processing.

\begin{table}[t]
    \caption{Ablation study of the value error recovery (VER) module. Lexicon features are used in all settings.  ``None'' means no post-processing, ``Delete'' means simply deleting the triplets that are invalid according to the ontology, and ``VER'' means applying VER module. We report the average F$_1$-score(\%)/joint-accuracy(\%) on the test set.}
    \label{tab:ablation_vr}
    \small
    \centering
    \begin{tabular}{l||c|c|c}
        \hline
        \multirow{2}{*}{\textbf{Models}} & \multicolumn{3}{c}{\textbf{post-processing settings}} \\
        \cline{2-4}
         & None & Delete & VER \\
        \hline\hline
        \textsc{HD} & 82.47/73.90 & 87.30/76.05 & 89.16/81.22 \\
        \textsc{Focus} & 88.75/81.81 & 91.10/83.34 & 91.46/84.31\\
        \textsc{DA-Align} & 88.53/82.12 & 91.25/83.59 & 91.69/84.74 \\
        \hline
        Proposed & \textbf{89.61}/\textbf{83.33} & \textbf{91.68}/\textbf{84.17} & \textbf{92.28}/\textbf{85.83} \\
        \hline
    \end{tabular}
\end{table}

\subsubsection{Ablation Study of the Training Procedure}

Table \ref{tab:ablation_training} considers (1) whether to pre-train the slot tagging model and (2) whether to utilize manual transcriptions in RL-training. Results indicate that pre-training using transcriptions helps to bootstrap the RL-training, and introducing transcriptions in RL-training stabilizes the training. Furthermore, the average performance decreases dramatically without these two procedures, which shows that the RL-training gets stuck in local optimum without any experiences about slot tagging from $\mathcal{D}_{tscp}$.

% \begin{table}[htbp]
%     \caption{F-score(\%)/joint-acc(\%) are evaluated on the test set and averaged over four domains. ``PT'' means pre-training as mentioned in section \ref{subsec:train_proc}, and ``Tscp'' means $\mathcal{D}_{tscp}$ are utilized in the RL-training stage. In terms of rewards, $R=R_\text{tscp}+R_\text{hyp}$.}
%     \label{tab:ablation}
%     \small
%     \centering
%     \begin{tabular}{ccc|c}
%         \hline
%         PT & Tscp & Reward & \textbf{avg.} \\
%         \hline\hline
%         \checkmark & \checkmark & $R$ & \textbf{91.41}/\textbf{84.09} \\
%         \checkmark & & $R$ &  91.18/83.65\\
%         & \checkmark & $R$ &  90.94/83.37\\
%         & & $R$ & 52.62/37.47\\
%         \hline
%         \checkmark & \checkmark & $R_\text{triple}$ & 91.29/83.85 \\
%         \checkmark & \checkmark & $R_\text{sent}$ & 91.36/83.95  \\
%         \hline
%     \end{tabular}

% \end{table}

\begin{table}[t]
    \caption{Ablation study of the training procedure for our proposed method. Lexicon features are utilized. Average F$_1$-score(\%)/joint-accuracy(\%) are reported.}
    \label{tab:ablation_training}
    \small
    \centering
    \begin{tabular}{c|c||c}
        \hline
        \textbf{Pre-training} on $\mathcal{D}_{tscp}$ & \textbf{Exploiting $\mathcal{D}_{tscp}$ in RL-training} & \textbf{avg.} \\
        \hline\hline
        \cmark & \cmark & \textbf{92.28}/\textbf{85.83} \\
        \hline
        \xmark & \cmark & 91.89/85.12\\
        \hline
        \cmark & \xmark & 91.87/85.01 \\
        \hline
        \xmark & \xmark & 43.06/34.69 \\
        \hline
    \end{tabular}

\end{table}

\subsection{Analysis}

\begin{figure}[t]
    \centering
    \includegraphics*[width=0.95\textwidth]{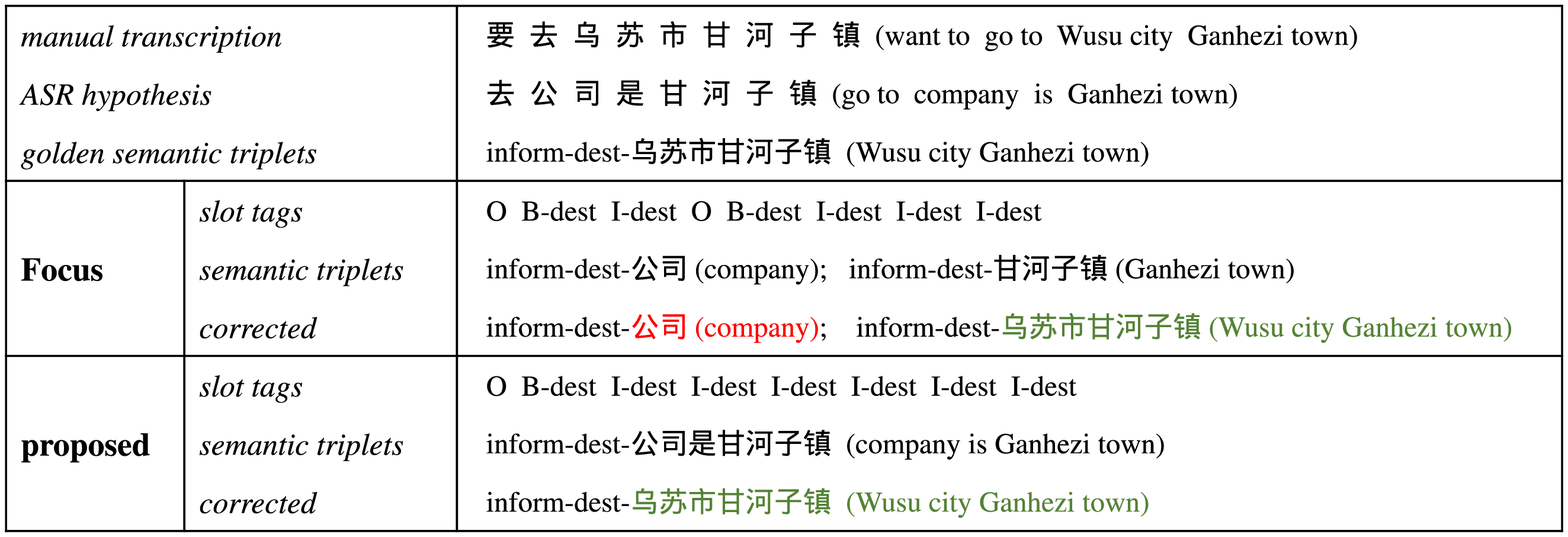}
    \caption{An example of slot tagging and value error recovery in \emph{map} domain. Note that during evaluation, only the ASR hypothesis is available as input.} % Wrong predictions are marked in red, while correct ones are in green.
    \label{fig:err_case}
\end{figure}

\subsubsection{Comparison with Baselines}

Traditional slot tagging models are supervised by BIO-tags, so ASR hypotheses without BIO-tag annotations cannot be used. In our method, the value error recovery module is applied to the output of the slot tagging model and provides feedback on the prediction. The feedback is utilized as a reward signal of RL-based training to finetune the slot tagging model. We give an example to illustrate how slot tagging benefits from VER guided training in Figure \ref{fig:err_case}. The baseline model recognizes two slot chunks ``公司~(company)'' and ``甘河子镇~(Ganhezi town)'' separated by a special word ``是~(is)''. The latter value is corrected by the VER module, whereas the former value is retained because it is also available in the value candidates corresponding to the 
act-slot pair \emph{inform-dest}, resulting in an incorrect triplet. By introducing the VER module during training, the tagging model learns to produce outputs more suitable for the subsequent VER module. Although word like ``是~(is)'' is unlikely to appear in a destination name, the tagging model considers it as a part of the slot, showing the capability to delimit the range of slots softly.

% DA method
In the view of data used, both \textsc{DA} methods and our proposed method utilize ASR hypotheses during training, while they treat hypotheses in different ways. \textsc{DA-Gen} uses a pre-trained tagging model to produce pseudo labels for ASR hypotheses. Therefore, noisy data is included for training, which will have a negative impact. In our proposed method, we can train the slot tagging model on ASR hypotheses with unaligned labels (i.e., \emph{act(slot=value)} triplets in this paper) directly.

%ASR hypotheses are also labelled by a tagging model, but those pseudo data are applied to fine-tune the tagging model with sequence-level rewards. 

\subsubsection{Different Character Error Rate (CER)}

\begin{figure}[t]
    \centering
    \includegraphics*[width=0.95\textwidth]{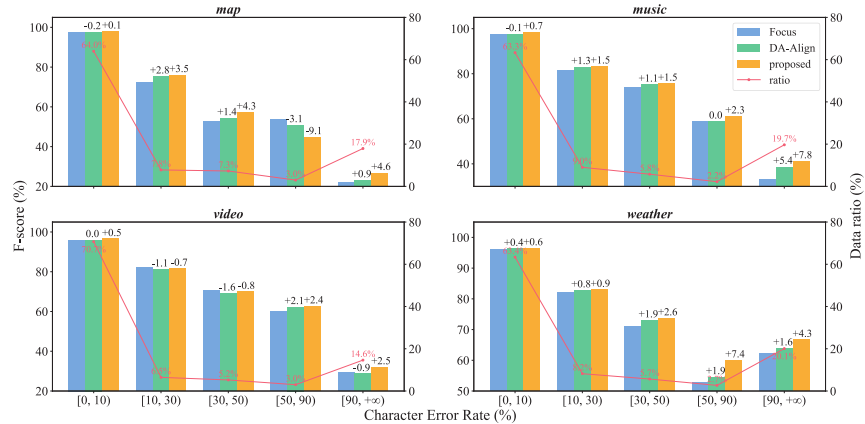}
    \caption{F$_1$-scores of our proposed model (with lexicon features) on the test set across four domains with various CER. The data ratio of each group is displayed in the form of line chart. The differences between the \textsc{Focus} baseline and the other two models are annotated on the figure for clarity.}
    \label{fig:cer_perf}
\end{figure}

To further investigate why our model achieves higher performance, we split the test set into various groups according to the character error rate (CER) of the ASR hypotheses, and compare our proposed method with the \textsc{Focus} and \textsc{DA-Align} baselines. The results in the four domains are presented in Figure \ref{fig:cer_perf}. With the increase of CER, the F$_1$-scores decline sharply. For utterances with low CERs (e.g., less than $10\%$), there is no significant difference (under $1\%$) between the baselines and our model. As the CER gets higher, our model can outperform the \textsc{Focus} and \textsc{DA-Align} by a larger margin. The improvements are particularly dramatic when the CER is higher than $90\%$. Note that exceptions happen occasionally, but in these cases, the amount of data is too small to draw a reliable conclusion. This finding further proves the robustness of our method against noisy ASR data.

\section{Related Work}
\label{sec:related_work}

% TODO

% slot tagging in SLU
SLU is often regarded as a sequence labelling problem modelled with Recurrent Neural Network (RNN)~\cite{mesnil2013investigation,yao2014spoken,liu2016attention,zhu2017encoder} and recent transformers~\cite{chen2019bert,qin-etal-2019-stack}. However, most of them assume that there are no ASR errors. To improve the robustness of SLU to ASR errors, previous works may apply input adaptations to reduce the gap between training and testing~\cite{zhu2018robust,schumann2018incorporating}, or correct predicted values by searching the most similar candidates in pronunciation~\cite{tan2019multi,wang2019transfer,li2019robust}. However, these two methods are not optimized jointly. There are other works to directly train the SLU model on ASR hypotheses~\cite{tur2013semantic,yang2015using,liu2020jointly}. However, these methods require qualified aligned data annotation on ASR hypotheses, which costs a lot. 

Except for the sequence labelling problem, SLU can also be directly considered as an unaligned task where outputs are semantic forms. In this view, unaligned annotations (semantic forms) can be transferred from manual transcriptions to ASR hypotheses straightforwardly. With unaligned data, SLU can be considered as a classification task~\cite{williams2014web} or a generative task~\cite{zhao2019hierarchical}. These methods do not require word-aligned labels but may lose generalization capability to unseen samples, which is confirmed by the \textsc{HD} baseline in our experiments. 

%Previous works handled these problems. Various types of hypotheses with richer information than ASR 1-best are introduced during training, including N-best lists~\cite{robichaud2014hypotheses,khan2015hypotheses}, word lattices~\cite{vsvec2015word} and word confusion networks~\cite{hakkani2006beyond,henderson2012discriminative,shivakumar2019confusion2vec}, while we only consider 1-best in this paper. A simple and straightforward solution is to annotate the ASR hypotheses at word level, but this is a time-consuming task, and hypotheses may vary when ASR system changes. 
%However, this post-processing module is usually fixed and independent of the SLU model, and the performance depends highly on the types of SLU errors.

% input adaptation

% RL
\section{Conclusion}

In this paper, we propose a robust SLU framework with a slot tagging model and value error recovery module. The value error recovery is utilized to guide the adaptive training of the slot tagging model on ASR hypotheses with reinforcement learning. Extensive experiments confirm that our model is more robust to ASR errors than the baselines.

\subsubsection{Acknowledgements.} We thank the anonymous reviewers for their thoughtful comments. This work has been supported by the National Key Research and Development Program of China (Grant No. 2017YFB1002102) and Shanghai Jiao Tong University Scientific and Technological Innovation Funds (YG2020YQ01).

\bibliographystyle{splncs04}
\bibliography{mybib}

\begin{thebibliography}{10}
\providecommand{\url}[1]{\texttt{#1}}
\providecommand{\urlprefix}{URL }
\providecommand{\doi}[1]{https://doi.org/#1}

\bibitem{chen2019bert}
Chen, Q., Zhuo, Z., Wang, W.: {BERT} for joint intent classification and slot
  filling. arXiv preprint arXiv:1902.10909  (2019)

\bibitem{goo2018slot}
Goo, C.W., Gao, G., Hsu, Y.K., Huo, C.L., Chen, T.C., Hsu, K.W., Chen, Y.N.:
  Slot-gated modeling for joint slot filling and intent prediction. In: NAACL
  (2018)

\bibitem{kingma2014adam}
Kingma, D.P., Ba, J.: Adam: A method for stochastic optimization. arXiv
  preprint arXiv:1412.6980  (2014)

\bibitem{li2019robust}
Li, H., Liu, C., Zhu, S., Yu, K.: Robust spoken language understanding with
  acoustic and domain knowledge. In: ICMI. pp. 531--535 (2019)

\bibitem{liu2016attention}
Liu, B., Lane, I.: Attention-based recurrent neural network models for joint
  intent detection and slot filling. In: INTERSPEECH. pp. 685--689 (2016)

\bibitem{liu2020jointly}
Liu, C., Zhu, S., Zhao, Z., Cao, R., Chen, L., Yu, K.: Jointly encoding word
  confusion network and dialogue context with bert for spoken language
  understanding. arXiv preprint arXiv:2005.11640  (2020)

\bibitem{mesnil2013investigation}
Mesnil, G., He, X., Deng, L., Bengio, Y.: Investigation of
  recurrent-neural-network architectures and learning methods for spoken
  language understanding. In: INTERSPEECH. pp. 3771--3775 (2013)

\bibitem{qin-etal-2019-stack}
Qin, L., Che, W., Li, Y., Wen, H., Liu, T.: A stack-propagation framework with
  token-level intent detection for spoken language understanding. In: Proc.
  EMNLP-IJCNLP. pp. 2078--2087 (2019)

\bibitem{schumann2018incorporating}
Schumann, R., Angkititrakul, P.: Incorporating {ASR} errors with
  attention-based, jointly trained {RNN} for intent detection and slot filling.
  In: ICASSP (2018)

\bibitem{sutton2000policy}
Sutton, R.S., McAllester, D.A., Singh, S.P., Mansour, Y.: Policy gradient
  methods for reinforcement learning with function approximation. In: NeurIPS
  (2000)

\bibitem{tan2019multi}
Tan, C., Ling, Z.: Multi-classification model for spoken language
  understanding. In: ICMI. pp. 526--530 (2019)

\bibitem{tur2013semantic}
T{\"u}r, G., Deoras, A., Hakkani-T{\"u}r, D.: Semantic parsing using word
  confusion networks with conditional random fields. In: INTERSPEECH. pp.
  2579--2583 (2013)

\bibitem{wang2019transfer}
Wang, X., Tang, C., Zhao, X., Li, X., Jin, Z., Zheng, D., Zhao, T.: Transfer
  learning methods for spoken language understanding. In: ICMI. pp. 510--515
  (2019)

\bibitem{williams2014web}
Williams, J.D.: Web-style ranking and {SLU} combination for dialog state
  tracking. In: SIGDIAL. pp. 282--291 (2014)

\bibitem{yang2015using}
Yang, X., Liu, J.: Using word confusion networks for slot filling in spoken
  language understanding. In: INTERSPEECH. pp. 1353--1357 (2015)

\bibitem{yao2014spoken}
Yao, K., Peng, B., Zhang, Y., Yu, D., Zweig, G., Shi, Y.: Spoken language
  understanding using long short-term memory neural networks. In: SLT (2014)

\bibitem{zhao2019hierarchical}
Zhao, Z., Zhu, S., Yu, K.: A hierarchical decoding model for spoken language
  understanding from unaligned data. In: ICASSP. pp. 7305--7309 (2019)

\bibitem{zhu2018robust}
Zhu, S., Lan, O., Yu, K.: Robust spoken language understanding with
  unsupervised {ASR}-error adaptation. In: ICASSP. pp. 6179--6183 (2018)

\bibitem{zhu2017encoder}
Zhu, S., Yu, K.: Encoder-decoder with focus-mechanism for sequence labelling
  based spoken language understanding. In: ICASSP. pp. 5675--5679 (2017)

\bibitem{zhu2019catslu}
Zhu, S., Zhao, Z., Zhao, T., Zong, C., Yu, K.: {CATSLU}: The 1st chinese
  audio-textual spoken language understanding challenge. In: ICMI. pp. 521--525
  (2019)

\end{thebibliography}

\end{CJK*}
\end{document}